\pdfoutput=1

\documentclass[11pt]{article}

\usepackage[]{EMNLP2022}

\usepackage{times}
\usepackage{latexsym}
\usepackage{url}
\usepackage{booktabs}
\usepackage[T1]{fontenc}
\usepackage{graphicx}
\usepackage[colorinlistoftodos]{todonotes}

\usepackage[T1]{fontenc}

\usepackage[utf8]{inputenc}

\usepackage{microtype}

\usepackage{inconsolata}

\usepackage{titlesec}
\titlespacing*{\paragraph}{\parindent}{0ex}{1ex}

\definecolor{Orange}{RGB}{255,140,0}
 
\newcommand{\figref}[1]{Figure~\ref{#1}}
\newcommand{\tabref}[1]{Table~\ref{#1}}
\newcommand{\secref}[1]{Section~\ref{#1}}
\newcommand{\Secref}[1]{Section~\ref{#1}}
\newcommand{\Wiki}{Concadia}

\title{Concadia: Towards Image-Based Text Generation with a Purpose}

\author{%
  Elisa Kreiss$^{1}$ \\\And
  Fei Fang$^{2}$ \\\And
  Noah D.~Goodman$^{2,3}$ \\\And
  Christopher Potts$^{1}$ \AND
  \\[-4ex] $^1$Department of Linguistics\quad$^2$Department of Computer Science\quad$^3$Department of Psychology\\
  Stanford University \\
  Stanford, CA 94305 USA \\
  \texttt{\{ekreiss, feifang, ngoodman, cgpotts\}@stanford.edu}
}

\date{}

\begin{document}
\maketitle
\begin{abstract}
Current deep learning models often achieve excellent results on benchmark image-to-text datasets but fail to generate texts that are useful in practice. We argue that to close this gap, it is vital to distinguish \emph{descriptions} from \emph{captions} based on their distinct communicative roles. \emph{Descriptions} focus on visual features and are meant to replace an image (often to increase accessibility), whereas \emph{captions} appear alongside an image to supply additional information. To motivate this distinction and help people put it into practice, we introduce the publicly available Wikipedia-based dataset \textbf{\Wiki} consisting of 96,918 images with corresponding English-language descriptions, captions, and surrounding context. Using insights from \Wiki, models trained on it, and a preregistered human-subjects experiment with human- and model-generated texts, we characterize the commonalities and differences between descriptions and captions. In addition, we show that, for generating both descriptions and captions, it is useful to augment image-to-text models with representations of the textual context in which the image appeared.
\end{abstract}

\section{Introduction}

Image-based natural language generation (NLG) has great potential for productive applications in accessibility, image search, creative writing, and navigational instruction tasks, among other areas. However, current systems fall short of realizing this potential even though they can generally produce fluent, truthful texts in a wide range of scenarios \citep[e.g.,][]{dognin2022image,guinness2018caption,gurari2020captioning}. In this paper, we identify and address one of the fundamental limitations holding these systems back: insufficient attention to the \emph{communicative purpose} of the text being generated.

\begin{figure}[tp]
  \centering
  \includegraphics[width=.42\textwidth]{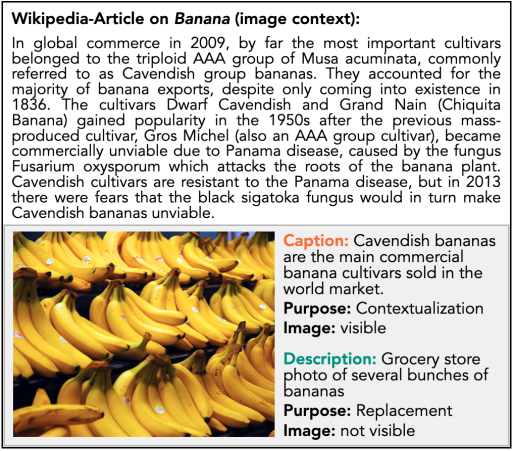}
  \caption{An example from our Wikipedia-based corpus \Wiki\ of an image from the article on bananas with two associated texts: the description as provided in the image's alt text, and the caption as displayed below the image in the article. For captions, the image content is presupposed whereas descriptions aim to stand in for the image. While grounded in the same image, description and caption convey vastly different information.} 
  \label{fig:data-sample-long}
\end{figure}

We focus on the high-level distinction between \emph{descriptions} and \emph{captions}. Descriptions are created for the purpose of replacing an image in context, most notably used to make an image accessible to users who can't see them. Captions, in contrast, presuppose access to the image and seek to provide additional information. For example, in \figref{fig:data-sample-long}, the description summarizes the scene and describes the features that are clearly visible in the image, whereas the caption provides almost no visual details and instead highlights the relevance of this image to the article. Both texts are grounded in the same image and have clear purposes, but neither can play the role of the other.

To enable in-depth exploration of the description/caption distinction, we introduce \textbf{\Wiki}, a Wikipedia-based corpus of 96,918 images with associated English-language captions, alt text descriptions, and accompanying context from the respective Wikipedia article. \Wiki\ provides the opportunity to directly compare descriptions and captions for shared images, and it can be used to develop image-based NLG systems that serve these distinct communicative purposes as well.

To further substantiate the description/caption distinction, we report on a linguistic analysis of the commonalities and differences between these two kinds of text in Wikipedia (\secref{sec:dataset}). We then develop and assess image-based NLG systems using Transformer- and LSTM-based architectures. Two clear lessons emerge from these modeling efforts. First, description generation is easier than caption generation, as one might expect given the tighter relationship between descriptions and images. Second, it is highly beneficial to include representations of the textual context in which an image appeared. This finding aligns with much recent work showing that blind and low-vision (BLV) users value image descriptions that tie the image to its surrounding context \citep{stangl2021going,stangl2020person,muehlbradt2022what}. Similarly, \citet{biten2019gooda} identify benefits of using context in the caption domain, for instance, for inferring names of people visible in an image.
In the example in \figref{fig:data-sample-long}, both the caption and description reflect the context (commercial aspects of bananas), and the caption additionally connects to the type of banana discussed in the article. Most datasets for image-based NLG provide only image--text pairs, with no context, which may fundamentally limit their usefulness. A strength of \Wiki\ is that all the image--text pairs are given within their Wikipedia contexts.

Finally, we provide evidence from a preregistered human-subjects experiment (\secref{sec:humanexp}). Our results further reinforce the observation that descriptions and captions serve distinct communicative purposes. In addition, the results hold not only for the descriptions and captions from \Wiki\ but extend to our model-generated texts as well. This serves to further validate that systems trained to generate captions are inadequate for the task of generating descriptions, and vice versa, and that model evaluations need to be sensitive to communicative purpose. Together with the modeling results, this speaks to the importance of both communicative purpose and context when it comes to developing effective image-based NLG systems.

\section{Captions vs.~Descriptions in Prior Work}

In AI, the task of generating prose text from images is often referred to as ``image captioning'', which glosses over distinctions of communicative purpose. The current section offers a synopsis of research, within and outside of AI, that is relevant to our proposed description/caption distinction and supports a more nuanced view of image-based NLG tasks.

The conceptual description/caption distinction we propose poses a terminological challenge. We use ``caption'' to mean ``image-supporting information expressed in text'', which we contrast with ``(alt) description'' to mean ``text intended to replace an image''. This is aligned with general popular usage and with the terminology in (photo-)journalism and graph accessibility, fields that conceptually already make a description/caption distinction. However, the term ``caption'' is now a loaded term in AI, and future work should be careful in considering the conceptual distinctions despite the challenges in terminology.

\subsection{Image-to-Text Relations}\label{sec:backgr-imgtextrel}

Most previous work has focused on characterizing the relation between images and the texts that occur alongside them. \citet{marsh2003taxonomy} summarize cross-disciplinary efforts, and propose a unifying taxonomy based on the different functions an image takes on in a text (e.g., from decorative through explanatory to interpretative). 

Work in communication sciences situates image--text relations based on their role in discourse \citep[e.g.,][]{martinec2005system}, and work in library sciences investigates the role that these relations play in efficient image indexing \citep[e.g.,][]{jaimes1999conceptual, shatford1986analyzing}.

More recently, the relations of images and visually co-occurring texts have been explored in AI, specifically for news \cite{otto2020characterization, oostdijk2020connection}, advertising \cite{otto2020characterization, zhang2018equal}, and social media \cite{kruk2019integrating, hodosh2013framing}.
\citet{alikhani2020cross} introduce an annotation protocol to investigate coherence relations between images and associated texts. They analyze to what extent texts from a variety of resources address what's \emph{visible}, \emph{subjective}, representing an \emph{action}, telling a \emph{story}, or containing \emph{meta} information. 
Our description vs.~caption distinction is focused on communicative purpose and can cross-cut these categories. 
For instance, evidence from BLV users suggests that a description intended to replace an image of politicians in a news article should include visible features that describe the scene, but also meta information that names the politicians and the actions they're engaging in \cite{stangl2021going}. 

A majority of the previously proposed categorizations 
are based on analyses of the text content, and most generally
assume that images and texts visually co-occur. The description vs.~caption distinction is a higher-level categorization that is based on whether the image content is presumed to be visible in the first place. This is motivated by communicative principles: when the image is visible, texts that simply describe the image would be redundant
, and instead those texts need to serve distinct communicative purposes 
\citep{kruk2019integrating}. 
\citet{hodosh2013framing} follow similar reasoning to explain why crowd-sourced image--text datasets at the time contained high rates of information that can't be directly extracted from the image. 
This distinction is particularly pronounced in the area of image accessibility. For instance, recent studies on the accessibility of computational research papers explicitly distinguish figure captions and figure descriptions when assessing a paper's accessibility \citep{williams2022toward,chintalapati2022dataset}. 

\subsection{Image-Based Text Generation Datasets}\label{sec:backgr-datasets}

We can loosely categorize existing datasets according to communicative purpose. \emph{Caption datasets} contain texts that are useful when appearing alongside images, mainly to contextualize them. \emph{Description datasets} contain texts that are intended to replace images. Some datasets are difficult to decisively place due to uncertainty about the nature of the texts or major postprocessing of the data.

\subsubsection{Description Datasets}

The first group of datasets were created by presenting sighted participants with images and the task to describe them. Presenting the images out of context restricts the text to information that can be extracted from the image itself. We therefore consider these to be \emph{description} datasets. 
Examples include MS-COCO \citep{lin2014microsoft,chen2015microsoft}, Flickr8k/30k \citep{young2014image,hodosh2013framing}, and more recently VizWizCaptions \citep{gurari2020captioning} and TextCaps \citep{sidorov2020textcaps}. 
Localized Narratives \citep{pont-tuset2020connecting} and DIDEC \citep{vanmiltenburg2018didec} are similarly constructed, except that the descriptions are spoken and the datasets further contain real-time data aligning regions of the image with the description.\footnote{See \citealt{vanmiltenburg2018varying} for a comparison of spoken vs.~written image descriptions.}

Conceptual Captions \citep{sharma2018conceptual} and its extension Conceptual 12M \citep{changpinyo2021conceptual} represent a second group of description datasets. 
These datasets contain crawled images and their associated alt texts from the Web. A primary purpose of alt texts is to replace the images when they're absent, and they are what screen readers use for providing non-visual image access. These use-cases align with the assumption that the text stands in for the image, which makes them description datasets. 

None of these datasets contain information on the context an image is situated in. In the case of crowd-sourced descriptions, no context was provided, which means that the description content solely relies on cognitive priors and heuristics to determine what is relevant. In the crawled datasets, we can generally assume that the people writing alt texts were aware of the context the images appear in, suggesting that the descriptions are likely to be context-sensitive. However, the context itself is often not preserved in the datasets, meaning that the effect of context can't be further investigated or learned by a model.

\subsubsection{Caption Datasets}

Some recent datasets focus on texts expressing information that can't be extracted from the image alone. The news domain is a prominent example \citep{biten2019gooda,ramisa2017breakingnews}, since the texts associated with images in newspapers generally presuppose that the user can see the image. Thus, we consider these to be \emph{caption} datasets. 
These datasets generally do contain contextual information in the form of the articles the images appear in.

\newcommand{\capdescnum}[2]{%
    \begin{tabular}[c]{@{}r@{}}caption: #1\\ description: #2\end{tabular}}

\begin{table*}[tp]
  \small
  \centering
  \begin{tabular}{l @{\hspace{14pt}} *{5}{r} }
    \toprule
    \textbf{split} & \textbf{datapoints} & \textbf{unique articles} & \textbf{avg length (words)}                                                 & \textbf{avg word length}                                                  & \textbf{vocab size}                                                           \\ \midrule
    train          & 77,534              & 31,240                   & \capdescnum{12.79}{14.53} & \capdescnum{6.07}{5.69} & \capdescnum{82,745}{54,893} \\[3ex]
    dev            & 9,693               & 4,952                    & \capdescnum{12.91}{14.57} & \capdescnum{6.10}{5.70} & \capdescnum{24,063}{16,800} \\[3ex]
    test           & 9,691               & 4,951                    & \capdescnum{12.67}{14.29} & \capdescnum{6.10}{5.71} & \capdescnum{23,650}{16,610} 
    \\[3ex]
    all            & 96,918              & 41,143                   & \capdescnum{12.79}{14.51} & \capdescnum{6.07}{5.69} & \capdescnum{96,908}{63,884} \\ 
    \bottomrule
  \end{tabular}
  \caption{Overview of our Wikipedia-based caption and description dataset \Wiki. Captions are on average shorter than descriptions but contain longer words. Captions make use of a larger vocabulary.} 
  \label{tab:corpus-overview}
\end{table*}

\subsubsection{Other Datasets}

Im2Text is a dataset crawled from Flickr, containing 1 million images and the user-generated text appearing alongside it. In this case, the purpose of the text is difficult to define since this same text is used for image search as well as providing information that might be read alongside the image. Recognizing this challenge, the dataset was postprocessed to make the text appear more descriptive. However, subsequent work has suggested that the dataset still contains a large number of non-descriptive texts \citep{hodosh2013framing}, making it difficult to decisively place.

While often not summarized in a single static dataset, Twitter data is also used for image-based text generation \citep[e.g.,][]{hessel2021clipscore}. The Twitter API offers access to the image as well as the tweet and, if present, a user-written image description to make it non-visually accessible. The tweet assumes that a user has access to the posted image which aligns it with the captioning purpose. The accessibility text seeks to substitute it, aligning with the description purpose.

The WIT dataset contains crawled images with their corresponding captions (which are called \emph{reference descriptions} in WIT), alt texts, and the surrounding paragraph from Wikipedia \citep{srinivasan2021wit}.\footnote{They also extract the image's \emph{Attribution Description}, which is a context-independent description of the image from Wikimedia, used for image search.} WIT provides a rich multilingual resource for images and associated texts in context. While it contains captions as well as alt descriptions for each image within a context, the authors note that the descriptions are rare and often contain only the filename. Perhaps for that reason, the subsequent analyses and experiments were not performed on the alt description data. \Wiki\ is specifically created for capturing the alt descriptions, and is controlled and filtered for their quality.

\section{The \Wiki\ Dataset}\label{sec:dataset}

We introduce \textbf{\Wiki}, \textbf{Con}textualized \textbf{c}aptions and \textbf{a}lt \textbf{d}escriptions from Wikiped\textbf{ia}, a corpus extracted from Wikipedia consisting of images with their naturally occurring alt descriptions, captions, and surrounding context. \Wiki\ allows for a direct comparison of descriptions and captions for a specific image in a specific context. In this section, we characterize the distribution of captions and descriptions on all of Wikipedia, describe the \Wiki\ dataset generation process, and provide a qualitative comparison on the similarities and differences of the captions and descriptions in \Wiki. Appendix~\ref{app:concadia} provides implementation details and further illustrations.

\subsection{Descriptions and Captions on Wikipedia}\label{sec:caption-description}

Descriptions and captions fulfill distinct communicative purposes that are already apparent in their distribution. Where present, captions are accessible for all readers, often printed below their respective image. Image descriptions are harder to access, since they're usually defined in the image's alt tag. Although hidden to most users, alt descriptions are essential for making images accessible to users who can't see them. The fact that alt tags are hidden by default likely decreases awareness and use of the field \cite{gleason2019it}, resulting in major accessibility challenges specifically for blind and low vision users.
The scarcity of alt descriptions online that could make these images nonvisually accessible is widely established, but existing studies have focused on social media \cite{morris2016most} and most frequently visited websites overall \cite{bigham2006webinsight,guinness2018caption}.
We expect to see a similar pattern on Wikipedia.

In 2020, English Wikipedia consisted of over 6 million articles with more than 4 million images.\footnote{This number is only an estimate, since confirmed numbers aren't available.} Based on 10,000 randomly sampled articles, our approximation suggests that while around 91\% of all images on Wikipedia are associated with a caption, only 6\% contain alt description tags. Moreover, this small subset includes alt descriptions that simply say \emph{alt text} or \emph{Image}. The number of informative alt descriptions is likely much lower.

\subsection{Dataset Curation}

For \Wiki, we extracted all images from Wikipedia that have captions as well as alt descriptions. Images were excluded where the picture wasn't publicly available at Wikimedia Commons, descriptions contained reference to the caption (e.g., \textit{refer to caption}), consisted of fillers (e.g., \textit{alt text}, \textit{text}), only consisted of the file name, or where any associated text was shorter than 3 characters. The extracted context paragraph is the paragraph following the image in the HTML, and is minimally 20 words long.

\tabref{tab:corpus-overview} provides basic corpus statistics. The final corpus consists of 96,918 images with descriptions, captions, and surrounding text from 41,143 articles. Additionally, we include a train/dev/test split for the data. Images that occur multiple times (but have potentially different associated texts), and images where caption and description are identical are sorted into the training set to ensure the highest quality in the validation and test sets. All other assignments are random, while ensuring that data from the same article are assigned to the same split. The dataset 
is publicly available.\footnote{\url{https://github.com/elisakreiss/concadia}}

\subsection{Caption/Description Similarities} \label{sec:semsim}

\begin{figure}[tp]
  \centering
  \includegraphics[width=.8\linewidth]{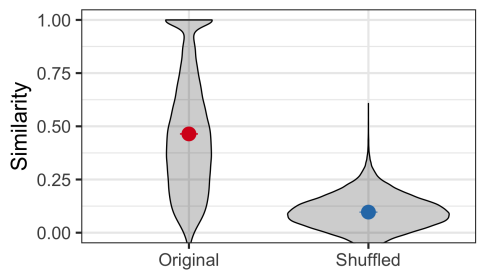}
  \caption{Cosine similarity between SBert embeddings of corresponding descriptions and captions (left, in red) and randomly sampled descriptions and captions (right, in blue) from \Wiki. The grey areas visualize the data distribution and the points mark average similarity with 95\% bootstrapped confidence intervals.} 
  \label{fig:sbert-sim}
\end{figure}

\Wiki\ provides the opportunity to directly compare descriptions and captions naturally written by Wikipedia users for a single image. If captions and descriptions show perfect similarity, there is no reason to assume a difference in communicative purpose, and captions could for instance simply replace the rarely present descriptions when they are absent. A complete dissimilarity would indicate a major effect of communicative purpose, and would suggest few commonalities that could be induced by the shared image and context. Crucially, if the content of captions and descriptions is \emph{partially} related, this would be compatible with the texts fulfilling distinct purposes while still being shaped by their shared image and context.

To investigate the semantic similarity between captions and descriptions, we computed the cosine similarity of the SBert embeddings \cite{reimers2019sentencebert} for all matching description/caption pairs (see \figref{fig:sbert-sim} in red).
Descriptions and captions are significantly more similar than would be assumed under a baseline where descriptions and captions are randomly paired (in blue; two-sample Kolmogorov-Smirnov test: $p<0.0001$), suggesting that they share item-specific content. (We replicate this using Jaccard distance of the raw strings, as presented in Appendix \ref{app:sim}.)

\begin{figure}[tp]
  \centering
  \includegraphics[width=.90\linewidth]{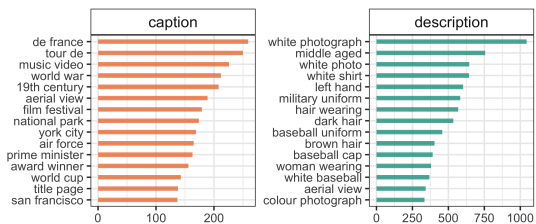}
  \caption{Most frequent bigrams, excluding stopwords (i.e., highly frequent words such as pronouns, prepositions, and forms of \textit{to be}).} 
  \label{fig:corpus-bigram}
\end{figure}

Having established that there is a semantic similarity between captions and descriptions, we turn to how they come apart. A quantitative analysis of the Parts-of-Speech (POS) shows distinct patterns for descriptions and captions. Adjectives and nouns are significantly more frequent in descriptions but captions contain significantly more proper nouns. This aligns with previous work on news captions \cite{biten2019gooda}, which found that proper nouns contribute 20\% of captions from New York Times articles while they're completely absent in datasets such as MS-COCO (a description dataset). This similarly replicates for the adjective and noun distributions \cite{biten2019gooda,ramisa2017breakingnews}. Intuitively, these patterns are already reflected in the most frequent bigrams on \Wiki\ (see \figref{fig:corpus-bigram}). Descriptions are dominated by descriptive attributes such as people's looks (e.g., \emph{white shirt}, \emph{baseball cap}, \emph{dark hair}), and meta information about the image (e.g., \emph{colour photograph}). The most frequent bigrams in captions are dominated by proper noun compounds such as \emph{San Francisco} or \emph{Tour de France}, as well as common places and times (e.g., \emph{national park}, \emph{19th century}). 

\section{Model Experiments}\label{sec:model}

Our investigation of the similarity between captions and descriptions in \secref{sec:semsim} indicates that they differ in principled ways but that they are not unrelated. This suggests that, while captions and descriptions can't stand in for each other, they can potentially inform each other's content when we automatically generate them. We further motivated that there is evidence that both description and caption generation would benefit from integrating the larger context where the image occurred. 
We now pursue these hypotheses in the context of large-scale caption and description generation.

\subsection{Model Architectures}

Previous work in image-based NLG has shown promising results from using both recurrent and Transformer-based models to generate text. To investigate the importance of a description vs.~caption distinction and the relevance of context, we compare three model variants. Depending on the condition, the models are trained on predicting description or caption labels and receive as context captions, descriptions, the broader context, or nothing. We provide further details on the models and training in Appendix \ref{app:models}.

\paragraph{ResNet-LSTM} generates a label token-by-token, using an LSTM decoder \cite{hochreiter1997long}. The input to the LSTM is a concatenated representation of the image using a pretrained ResNet \cite{he2016deep}, and the context using BERT embeddings \cite{devlin2019berta}. We took inspiration from the importance of attention mechanisms \citep{vaswani2017attention} on the input image as used in state-of-the-art ``captioning'' models \cite{xu2015show,lu2017knowing,anderson2018bottomup} and extended them to the context input. To ensure the same number of trainable parameters between the with-context vs.~no-context models, the no-context models received all-ones vectors in place of context embeddings.

\paragraph{DenseNet-LSTM} follows the same structure as the ResNet-LSTM but leverages pretrained DenseNet features instead 
\cite{deng2020image,hossain2021text}.

\paragraph{OSCAR(VinVL)} is a pretrained Transformer-based vision-language model that achieves state-of-the-art results on many tasks via finetuning \cite{zhang2021vinvl}. For each image, the VinVL pretrained visual feature extractor provides visual features, i.e., vector embeddings of object-enclosing regions, and object tags in text. To obtain a unified representation of the image--text pair, the OSCAR model concatenates the BERT embeddings \cite{devlin2019berta} of the text and object tags with the visual features \cite{li2020oscar}. We supply context by appending it to the object tags before the BERT encoding. Since BERT can handle variable-length inputs, the trainable parameters remain constant regardless of whether context is incorporated.

\subsection{Evaluation Metrics}\label{sec:model-eval}

Since \Wiki\ is extracted from naturally occurring data, each image is only associated with a single ground-truth caption and description. Currently established evaluations of text generated from images rely on multiple references to reliably estimate performance. As suggested in previous work \cite{biten2019gooda}, we consider CIDEr \cite{vedantam2015cider} to be most appropriate for our setting. First, CIDEr has been shown to outperform BLEU \cite{papineni2002bleu} and ROUGE \cite{lin2004rouge} specifically when reference sentences are sparse \cite{vedantam2015cider,anderson2016spice}. Second, while METEOR \cite{denkowski2014meteor} and SPICE \cite{anderson2016spice} achieve slightly higher alignment with human judgments on some datasets with few references, their use of soft-similarity and lemmatization is not well-defined for proper names \cite{biten2019gooda}, which make up more than 26\% of our caption data and more than 15\% of our description data (\secref{sec:semsim}).

\subsection{Experiment 1}

In our first experiment, we probe how the distinction between the tasks of description and caption generation affects model performance.

\paragraph{Predictions}
If descriptions are meant to replace an image while captions are meant to complement it, we expect models to perform better at generating descriptions from image input alone, as compared to captions. In Experiment~1, we test whether this prediction is borne out in the three models. A high performance gap between the two conditions would further highlight the necessity of differentiating those tasks to address their distinct challenges.

\paragraph{Results}
\tabref{tab:modelresults} shows the performance on the test set for description and caption generation. Test set evaluation was performed using beam search ($n=5$) for all models. As predicted, all models achieve higher performance on the description data than the caption data. This suggests that the information from the image alone is more helpful to description generation, suggesting a closer connection between the two. 
Experiments using shuffled versions of the dataset further show that the performance gap can't be explained simply by the smaller vocabulary in descriptions, the higher proportion of proper nouns in captions, or the higher n-gram overlap across descriptions (see \Secref{app:model_perfgap}). 

\subsection{Experiment 2}

\begin{table}[tp]
\centering
\small
\begin{tabular}{@{} l@{ \ }l@{ \ }l@{ \ }l@{ \ }l@{ \ }@{}}\toprule
  Label & Context & RN-LSTM & DN-LSTM & OSC.VinVL \\ 
  \midrule 
Descr. & None & 0.16 (0.00) & 0.21 (0.00) & 0.21 (0.00) \\
Caption & None  & 0.07 (0.00) & 0.10 (0.01) & 0.10 (0.01) \\
\midrule
Descr. & Rand. Par. & 0.14 & 0.19 & 0.22 \\
Caption & Rand. Par. & 0.06 & 0.09 & 0.10 \\
\midrule
Descr. & Paragr. & 0.20 (0.00) & 0.23 (0.01) & 0.32 (0.01)  \\
Caption & Paragr. & 0.13 (0.00) & 0.15 (0.00) & 0.45 (0.00)  \\ 
\midrule
Descr. & Caption & 0.27 (0.01) & 0.28 (0.00) & 1.18 (0.01) \\
Caption & Descr. & 0.17 (0.01) & 0.19 (0.00) & 1.14 (0.01) \\ \bottomrule
\end{tabular}
\caption{CIDEr scores for all models: ResNet-LSTM, DenseNet-LSTM and OSCAR(VinVL). Standard deviations (in brackets) are computed over 3 random seeds. Across models, description generation results in higher CIDEr scores and adding contexts boosts performance.}
\label{tab:modelresults}
\end{table}

We have seen that descriptions and captions play different roles and pose different modeling challenges. Can these two kinds of text mutually reinforce each other in the context of automatic text generation? 
We investigate this question by treating one as the context that can inform generation for the other. For a second notion of context, we also test whether the closely occurring paragraph can help the models' performances. A positive effect of context would point to the value of treating image-based NLG as a contextual problem.

\paragraph{Predictions}
We predict that, while distinct in purpose, descriptions and captions from the same image and context share enough information that they can benefit from one another. Part of this contextual relevance should also be captured when using the closely appearing paragraph. To ensure that not just language in general accounts for potential performance gains, we provide control conditions where the model receives paragraphs from other images. Improvements over these controls can then only be due to semantic relatedness between the context and the generated description/caption.

\paragraph{Results}
\tabref{tab:modelresults} shows that, where models are provided contextual representations, performance improves for both description and caption generation compared to the image-only input baselines. As predicted, this improvement is most pronounced when providing the respective caption or description. Even the closely appearing paragraph leads to increased CIDEr scores compared to the no-context baselines. Furthermore, this improvement goes beyond what models achieved when training on randomly assigned paragraphs, which suggests that they make use of item-specific information and points to the more general benefit of integrating the broader context an image appears in.

\subsection{Discussion}
\label{app:model_perfgap}

Across models, we find that they achieve higher CIDEr ratings on the description than the caption generation task (see \tabref{tab:modelresults}). In part this might be due to the smaller set of tokens present in descriptions (\tabref{tab:corpus-overview}), the higher proportion of proper nouns in captions (\figref{fig:poscomparison}), and the higher n-gram overlap across descriptions (\figref{fig:corpus-bigram}). An additional potential explanation for the gap is that descriptions are more tightly connected to their image than captions, and are therefore easier to learn. 

The first set of factors are image-independent, i.e., they are simply properties of the language labels. If these components solely drive the higher performance on the description data, this gap should be reflected in a version of the dataset where the image-label mappings are shuffled. We conducted this experiment using the ResNet-LSTM model instead of the OSCAR(VinVL) model since the OSCAR(VinVL) model already carries biases from its pretraining. Although the shuffled description model achieves higher CIDEr scores than the shuffled caption model (0.017 vs.~0.016, respectively), it doesn't compare to the performance gain on the non-shuffled data (0.09). The tighter relationship of descriptions and their images predicts this gap and falls in line with the general description/caption purpose distinction.

Further support for the purpose of the description/caption directly affecting their predictability comes from the experiment where we provided the closely appearing paragraph as the context for the images (\tabref{tab:modelresults}). In the best-performing model, OSCAR(VinVL), caption generation performance surpasses description performance when the paragraph is provided. Furthermore, while caption performance doesn't surpass description generation performance in the LSTM models, providing the paragraph as context to the caption models results in a higher performance increase than for the description models. This is intuitive given the predicted connection of the descriptions/captions to the images and context. If the primary purpose of the caption is to connect the image to its context, the paragraph should be crucial for generating it. For the description, on the other hand, the image takes on a primary role, and adding the context should result in lesser gains. 
Taken together, the quantitative model results are in line with the description/caption purpose distinction, suggesting a closer alignment of the description with the image and the caption with the context.

In sum, we find that providing linguistic context has the potential to improve model-generated descriptions as well as captions. These results also have implications for applications. Our results suggest that models trained to generate descriptions, which are generally sparse, might benefit from receiving the more available caption information as additional input, providing a new potential path for addressing image accessibility challenges.

\section{Human Evaluation}\label{sec:humanexp}

In \secref{sec:semsim} and \secref{sec:model}, we provided initial evidence that descriptions and captions fulfill separate communicative purposes. We now present a human-subject experiment in which we directly test whether captions are more suitable than descriptions for providing information that can't be obtained from the image alone, and whether descriptions are more useful than captions for replacing an image.
Furthermore, we investigate whether models might capture these differences. Our experiment focuses on the Resnet-LSTM model, the weakest of our models, which should then provide a conservative assessment of how closely automatic generations align with human texts.

Our quantitative and qualitative hypotheses, along with our exclusion criteria and the analyses, were preregistered.\footnote{OSF preregistration link: \url{https://osf.io/qw9nr}.} Details on the recruitment and data exclusions are provided in Appendix \ref{app:humaneval}.

\paragraph{Materials} We sampled 300 images from the validation set in \Wiki. In the experiment, these could be paired with their original descriptions from \Wiki, their original captions, or the model-generated descriptions or captions, resulting in four text conditions for each image.

\begin{figure*}[tp]
  \centering
  \includegraphics[width=0.98\linewidth]{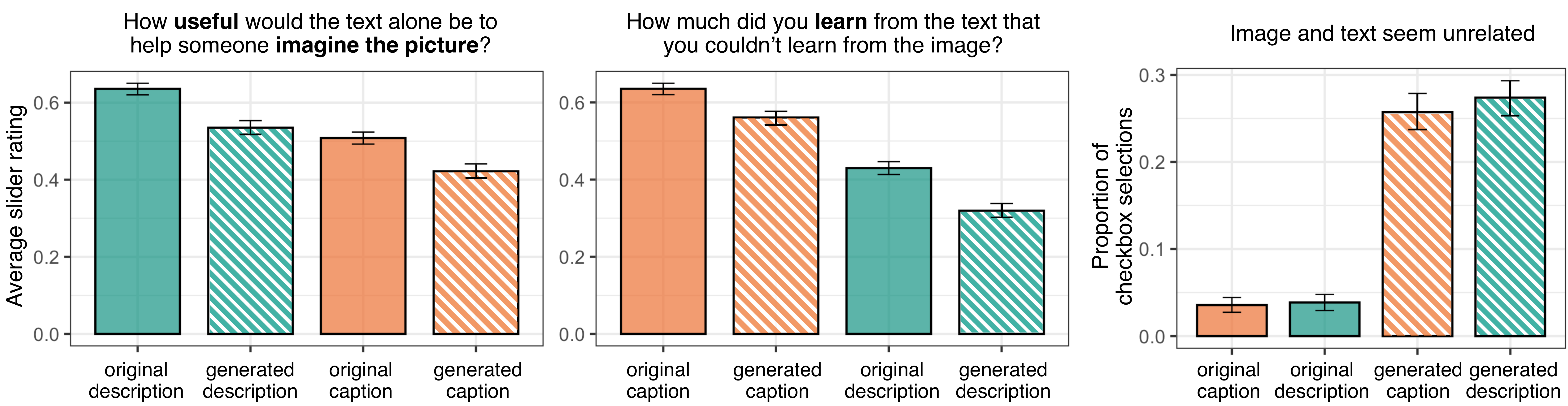}
  \caption{Results from the human subject experiments showing that descriptions and captions fulfill separate purposes and that these differences are picked up by models trained on them. Error bars indicate 95\% bootstrapped confidence intervals.} 
  \label{fig:humanexpresults}
\end{figure*}

\paragraph{Design} In each trial, an image was displayed next to one of the descriptions/captions sampled from the four text conditions, followed by two questions (\figref{fig:humanexpresults}). Participants assessed how useful the text would be to help someone imagine the picture (rated from \emph{not useful} to \emph{very useful}), and how much they learned from the text that they couldn't have learned from the image (rated from \emph{nothing} to \emph{a lot}). Each question was associated with a continuous scale initialized at zero and ranging to one. Participants could further opt out of responding to the questions by selecting a checkbox to indicate that image and text seemed to be unrelated.

\paragraph{Procedure} Each participant saw 32 unique images, and the corresponding texts were uniformly sampled from the four text conditions. Image selection and trial order were randomized and question order was randomized between participants.

\paragraph{Participants} We recruited 421 participants over Amazon’s Mechanical Turk, who were paid \$2.60 for participation with an average completion time of 12 minutes (\$13/hr). After preregistered exclusions, the analysis is based on 277 participants.

\subsection{Predictions}

Overall, we predicted that descriptions and captions fulfill different communicative purposes. Concretely, we made the following predictions for the original human-generated descriptions and captions in \Wiki:
\begin{description}\setlength{\itemsep}{0pt}
\item[H1] The original descriptions are judged more useful for imagining the picture than the original captions.
\item[H2] A reader receives more extra-image information from the original captions than from the original descriptions.
\end{description}

We further predicted that a model should be capable of reflecting these differences when trained on descriptions and captions separately. Predictions (H1) and (H2) should therefore also hold for model-generated descriptions and captions.

Finally, we qualitatively predicted that models would be strong enough to show the following: generated descriptions should be more similar to original descriptions than the original captions would be, and vice versa for generated captions.

\subsection{Results}

To test our quantitative predictions, we performed a Bayesian mixed effects regression analysis, where we predicted continuous slider ratings from the centered categorical text condition variable with random intercepts and slopes for each participant and image. 
We find strong evidence for all of our quantitative predictions and we also see the predicted qualitative patterns, as shown in \figref{fig:humanexpresults}.
The results lead us to two main conclusions. Firstly, there are multiple dimensions in which an image-text pair can be successful. Our results show that we can reverse which texts are rated as ``better'' depending on what goal we prioritize: the ability of to reconstruct the image from the text in its absence, or the amount of information learned from the text in its presence.
Secondly, models trained on caption and description data separately capture core aspects of these differences. Even with its limitations, the generated descriptions and captions provide more useful alternatives for their original counterparts than original captions and descriptions, respectively. 

Unsurprisingly, the generated text was more often rejected as being unrelated to the image than the original descriptions and captions were, which were only rejected in less than 5\% of all trials. These results underline the quality of \Wiki\ and shortcomings of the ResNet-LSTM model.

Overall, our results highlight the importance of specifying the purpose a text is intended to fulfill in the image-text generation domain -- both for dataset creation and model evaluation. The questions posed for evaluation have to be assessed against the purpose that they're intended to fulfill. 

\section{Conclusion}

We argue for a distinction between \emph{descriptions} (text intended to replace an image) and \emph{captions} (text intended to contextualize an image), and we provide evidence for their similarities and differences through linguistic analyses and human-subject evaluations. We introduce a corpus, \Wiki, consisting of images and their corresponding alt descriptions, captions, and context, and we show that \Wiki\ is a valuable resource for training NLG systems for both description and caption generation. The structure of \Wiki\ additionally allows us to show that both generation tasks benefit from contextual information. Beyond the conceptual relevance for advancing image-based NLG systems, this has practical implications specifically for image accessibility. Captions which often co-occur with images that lack an accessibility description could be a valuable resource for building more powerful description generation models.

\section*{Limitations and Ethics}

The proposed dataset \Wiki\ is crawled from Wikimedia Commons. Our understanding is that the images and associated texts are compliant with the policies of Wikimedia and Wikipedia. 

\Wiki\ is a subset of Wikipedia, filtered for images that had associated alt texts. Therefore, data might disproportionately be extracted from articles historically containing many images, such as the Tour de France or Award Shows, frequently visited articles where lacking image accessibility is more likely to be noticed, and from authors that are aware of the alt texts' uses, for instance, from accessibility-related Wikipedia articles. Additionally, \Wiki\ inherits biases generally present on Wikipedia (extraction date: December 2020). For instance as of February 2020, only 19\% of biographical articles were about women \cite{tripodi2021ms}, and it's likely that gender as well as ethnicity biases are perpetuated in \Wiki.
\Wiki\ is further constrained to English-language entries since this is the language we chose for our case study.
We provide a corresponding datasheet in our supplementary materials \cite{gebru2021datasheets}.

The image-based natural language generation models were trained to allow for comparisons between description and caption generation, and to investigate the role of context in both cases. The models are not intended for deployment of any kind since they were not tested for crucial dimensions such as their robustness, truthfulness, or output of potentially discriminatory language. The model outputs further need to be tested to determine the extent to which they actually capture the intended use cases. For instance, a description might be intended to serve an image accessibility purpose, but rigorous testing is required to determine the extent to which the data and associated models fulfill this goal for blind and low vision users.

The human subject experiment was conducted under an IRB protocol. Participants were paid \$2.60 with an average completion time of 12 minutes (\$13/hr).

\section*{Acknowledgements}

This work is supported in part by a grant from Google through the Stanford Institute for Human-Centered AI. We thank our experiment participants for their invaluable input. We are further grateful to Desmond Elliott, and Dan Jurafsky for their thorough feedback on previous drafts. We thank Rachit Dubey, all members of Stanford's CoCoLab, and Jessica Mankewitz for their insightful comments on this work, and to Mike Wu for sharing useful materials.

\bibliographystyle{acl_natbib}
\bibliography{Captioning}

\appendix

\section*{Appendix}

\section{Additional Details on \Wiki}\label{app:concadia}

\begin{figure}[b!]
  \centering
  \includegraphics[width=0.465\textwidth]{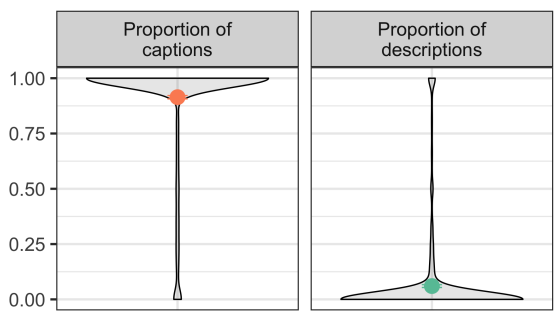}
  \caption{Proportion of images with captions vs.~descriptions on Wikipedia, as of December 2020. The grey areas visualize the data distribution and the points mark average proportion of captions/descriptions for an image with 95\% bootstrapped confidence intervals. Each datapoint is the average proportion for each of the 10,000 randomly sampled articles.} 
  \label{fig:wiki-sample}
\end{figure}

\begin{figure}[tp]
  \centering
  \includegraphics[width=0.9\linewidth]{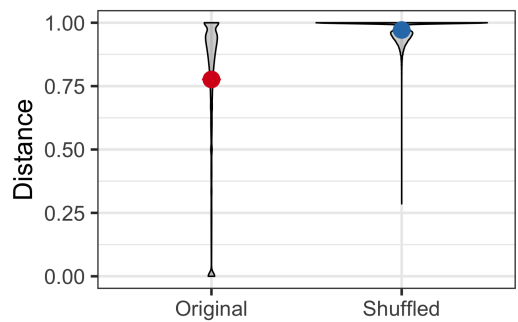}
  \caption{Jaccard distance between the originally matched captions and descriptions and shuffled caption-description pairs.} 
  \label{fig:jaccdist}
\end{figure}

\subsection{Datasheet}\label{app:datasheet}

A datasheet \citep{gebru2021datasheets} for \Wiki\ is included in our supplementary materials.

\subsection{Occurrence Frequency Estimates}
To estimate the sparsity of alt descriptions and captions for images on Wikipedia, we randomly sampled 10,000 articles, extracted the number of images, and counted how many of those contained captions and how many contained descriptions. We then took the average proportion of captions and descriptions per image for each article, which yields the data distribution displayed in grey in \figref{fig:wiki-sample}. 

\begin{figure}[b!]
  \centering
  \includegraphics[width=0.48\textwidth]{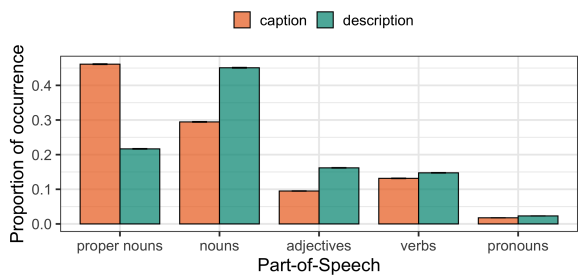}
  \caption{Aggregated proportions of Part-of-Speech frequencies for descriptions and captions with 95\% confidence intervals. The following POS tags constitute the categories on the x axis: proper nouns (NNP, NNPS), nouns (NN, NNS), adjectives (JJ, JJR, JJS), verbs (VB, VBD, VBG, VBN, VBP, VBZ), pronouns (PRP, PRP\$).} 
  \label{fig:poscomparison}
\end{figure}

\begin{figure*}
  \centering
  \includegraphics[width=1\textwidth]{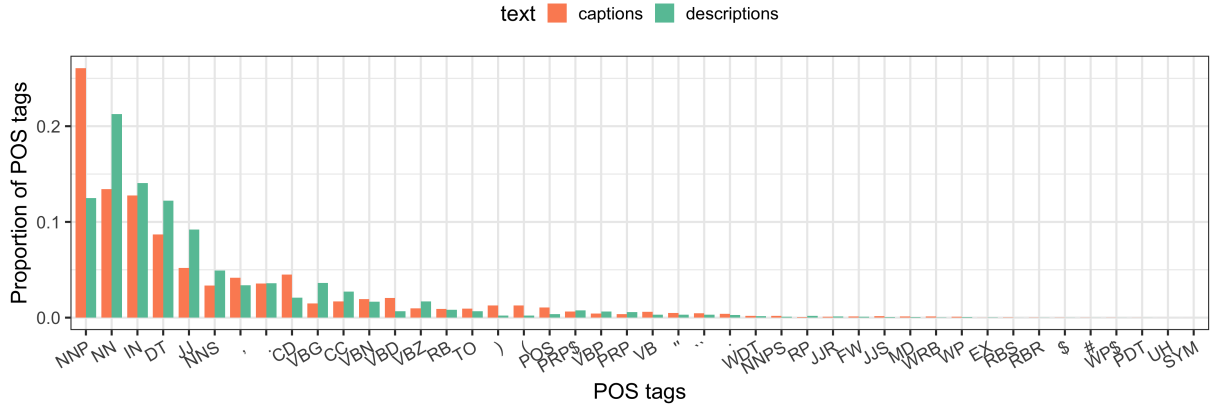}
  \caption{Proportion of part-of-speech tokens in descriptions vs.~captions from \Wiki.} 
  \label{fig:app-postags}
\end{figure*}

\subsection{Similarity of Captions and Descriptions}\label{app:sim}

In \secref{sec:semsim}, we demonstrate that captions and descriptions for the same image in \Wiki\ are semantically similar to each other using SBert embeddings. These results replicate when calculating similarity using Jaccard distance of the raw strings, a more form-based and superficial perspective on the caption/description comparison. \figref{fig:jaccdist} summarizes this analysis. The random pairings are significantly more distant from each other than their ordered counterparts (two-sample Kolmogorov-Smirnov test: $p<0.0001$). The results still support the observation of distinctness between captions and descriptions since the Jaccard distance in the ordered set is still high ($0.78$).

In \secref{sec:semsim}, we further present differences in occurrence frequencies of the syntactic categories \emph{proper nouns}, \emph{nouns}, and \emph{adjectives} for descriptions and captions, as illustrated in \figref{fig:poscomparison}. However, the Part-of-Speech (POS) tag analysis using the Python library NLTK \citep{bird2006nltk} allows for much more finegrained distinctions. \figref{fig:app-postags} contains all analyzed POS tags and shows the proportion at which they occur in captions and descriptions of the \Wiki\ corpus, revealing informative sub-patterns. For example, while there is a clear over-representation of adjectives in descriptions, this pattern qualitatively flips for superlatives. We speculate that superlatives are potentially more evaluative and therefore considered less appropriate in descriptions. There is also a clear over-representation of determiners in descriptions, which we attribute to the higher frequency of common nouns over proper names, since in English \textit{Jesse} doesn't need a determiner to make an utterance grammatical whereas \textit{person} most often does. Proportions are computed by dividing the number of occurrences of a POS tag by the number of all tokens in that specific text form. This allows for a direct comparison between descriptions and captions, even though captions contain more tokens overall. \tabref{tab:app-posexpl} contains an overview of all POS tags and their respective definition, as provided by NLTK \citep{bird2006nltk}. 

\begin{figure}[tp]
  \centering
  \includegraphics[width=0.49\textwidth]{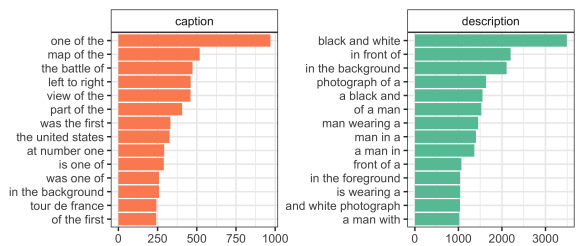}
  \caption{Most frequent trigrams in \Wiki\ for captions and descriptions.} 
  \label{fig:app-trigrams}
\end{figure}

\secref{sec:semsim} also addresses qualitative semantic differences between captions and descriptions. Due to space constraints, we only use the most frequent bigrams (without stopwords) to demonstrate these differences in the main paper. \figref{fig:app-trigrams} illustrates the most frequent trigram frequencies when including stopwords. Similarly to the bigrams in \figref{fig:corpus-bigram}, the most frequent trigrams in captions have instances of event descriptions and proper names (e.g., \textit{the battle of}, \textit{Tour de France}). For descriptions, there are again primarily descriptive attributes for people (e.g., \textit{man wearing a}, \textit{is wearing a}), and meta information about the image (e.g., \textit{(black) and white photograph}, \textit{in the foreground}).

\begin{figure}[tp]
  \centering
  \includegraphics[width=0.47\textwidth]{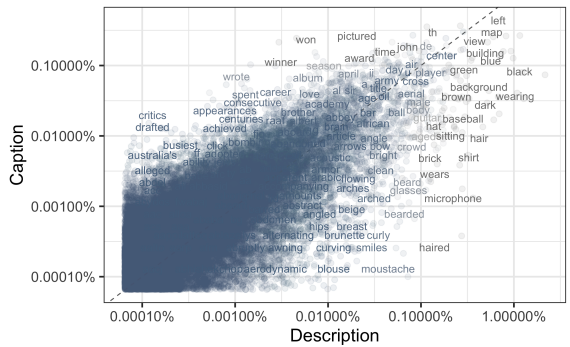}
  \caption{Correlation of word frequency between captions and descriptions.} 
  \label{fig:app-wordfrqcorr}
\end{figure}

A similar pattern arises when directly comparing which words occur more frequently in captions and descriptions, as shown in \figref{fig:app-wordfrqcorr}. Words that lie on the dashed line occur equally often in both text forms. Words below that line occur more frequently in descriptions and words above it occur more often in captions. Words that occur more frequently in descriptions focus on descriptive adjectives and nouns such as \textit{hat}, \textit{bearded}, or \textit{blouse}. Words that occur more frequently in captions rather refer to an event, e.g., \textit{award} or \textit{winner}, and use more evaluative language, e.g., \textit{achieved}.

Finally, a simple binary classifier based on the SBert embeddings of the descriptions and captions achieves an accuracy of 79.3\% (train set: 79.5\%, dev set: 79.5\%) providing further quantitative evidence suggesting that descriptions and captions carry distinguishable features.

These additional perspectives are in further support of the quantitative and qualitative analyses of the principled ways in which the content of captions and descriptions differs (see \secref{sec:semsim}).

\subsection{Illustrations}

In \figref{fig:app-concadiasamples}, we provide examples from \Wiki, illustrating a range of topics and image types: photographs, drawings and paintings.

\begin{figure*}
  \centering
  \includegraphics[width=1\textwidth]{figures/concadia_sample.pdf}
  \caption{Sample images with the corresponding descriptions, captions and excerpts of the context paragraphs from \Wiki.} 
  \label{fig:app-concadiasamples}
\end{figure*}

\section{Models}\label{app:models}

\subsection{ResNet-LSTM}

\begin{figure}[tp]
  \centering
  \includegraphics[width=1\linewidth]{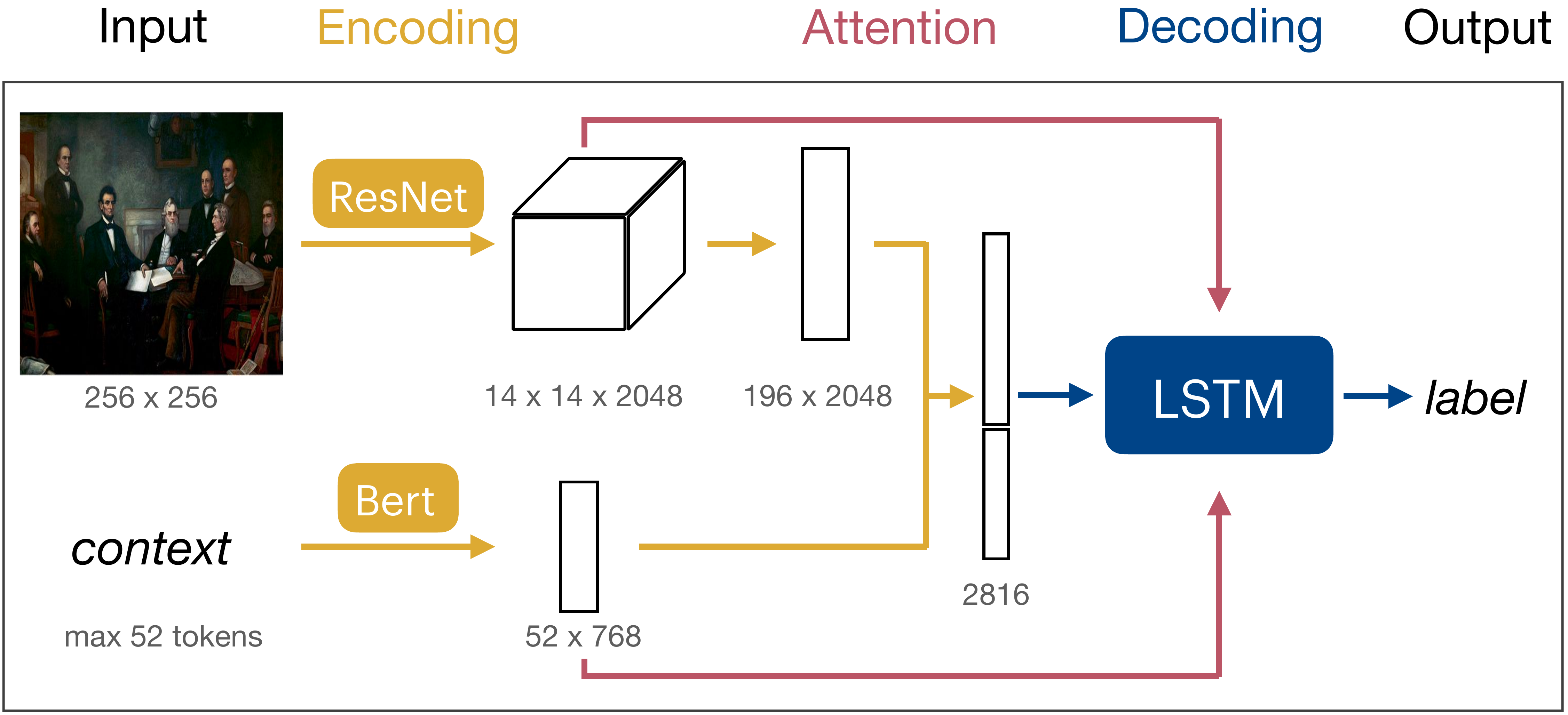}
  \caption{ResNet-LSTM model architecture. The LSTM decoder takes image as well as text input with attention on both representations. If no context is provided, the context representation contains ones only. In our experiments, the model receives the caption as context to produce descriptions, and vice versa.} 
  \label{fig:model-architecture}
\end{figure}

\subsubsection{Architecture}

\paragraph{Encoders} Two pretrained encoders create image and context representations which are input to an LSTM with attention. As an image encoder, we use ResNet-101 \cite{he2016deep}, pretrained on ImageNet.\footnote{We used the pretrained weights available with Pytorch's torchvision module version 0.8.2.}
For the context representation, we use pretrained BERT \cite{devlin2019berta}.\footnote{We used the pretrained weights available with the Python module pytorch-pretrained-bert version 0.6.2.} When predicting labels in absence of context, we induce an uninformative context representation by setting all values to 1. This way, we ensure that potential model performance improvements are not due to an increase of model complexity when adding context.
This is also one of the reasons why neither image nor context encoders were finetuned. To establish the importance of input components, e.g., context, we needed to fix the number of trainable parameters to avoid conflations from changes in model complexity. 

\paragraph{Decoder with Attention} The initial input to the LSTM hidden states of the decoder is a vector concatenating an image and a context representation. The LSTM decoder generates each token in a predicted label based on a previous token, the previous hidden state, and an attention vector. The attention vector is obtained by using soft attention \cite{xu2015show,bahdanau2015neural} separately parameterized on the image and the context and then concatenating the resulting representations.

\begin{table}
  \centering
  \setlength{\tabcolsep}{3pt}
  \begin{tabular}{@{} ll @{}}
    \toprule
    \textbf{POS tags} & \textbf{Explanation}                  \\ \midrule
    CC                          & Coordinating Conjunction              \\
    CD                          & Cardinal Digit                        \\
    DT/PDT                      & Determiner/Predeterminer                            \\
    EX                          & Existential There                     \\
    FW                          & Foreign Word                          \\
    IN                          & Preposition/Subordin. Conjunct. \\
    JJ/JJR/JJS                  & Adj./Comparative/Superlative \\
    LS                          & List Marker 1                         \\
    MD                          & Modal                                 \\
    NN/NNS                      & Noun Singular/Plural                        \\
    NNP/NNPS                   & Proper Noun Singular/Plural              \\
    
    POS                         & Possessive Ending                     \\
    PRP/PRP\$                         & Personal/Possessive Pronoun      \\
    RB/RBR/RBS                  & Adverb/Comparative/Superlative     \\
    RP                          & Particle                              \\
    TO                          & to                                    \\
    UH                          & Interjection                          \\
    VB                          & Verb, Base Form                       \\
    VBD/VBN                    & V, Past Tense/Past Participle                      \\
    VBG                         & V, Gerund or Present Participle       \\
    VBP/VPZ                  & V, Sg. Present, non-3rd/3rd P.         \\
    WDT/WRB                 & wh-determiner/wh-abverb                     \\
    WP/WP\$                        & wh-pronoun/possessive\\
  \bottomrule
  \end{tabular}
  \caption{Part-of-speech tokens from \figref{fig:app-postags}.} 
  \label{tab:app-posexpl}
\end{table}

\subsubsection{Implementation}\label{sec:modelimpl}

The ResNet-LSTM models were implemented in PyTorch \cite{paszke2019pytorch} using a codebase that has successfully replicated results from \citet{xu2015show}.\footnote{\url{https://github.com/sgrvinod/a-PyTorch-Tutorial-to-Image-Captioning}} 
Additional details on model training and optimization will be available in our code release.

\subsubsection{Hyperparameters and Training Details}\label{app-hyperparam}

We used the hyperparameters suggested in this codebase. The only exception is that we used a reduced batch size (32 instead of 80) since our models that include context have more parameters.

Overview of hyperparameters:

\begin{itemize}
    \itemsep0em 
    \item dimension of image encoder output: 2048 (predetermined by the pretrained ResNet-101 \citet{he2016deep} output size)
    \item dimension of context encoder output: 768 (predetermined by the pretrained Bert \citet{dognin2019adversarial} output size)
    \item dimension of word embeddings: 512
    \item dimension of attention linear layers: 512
    \item dimension of decoder RNN: 512
    \item dropout: 0.5
    \item loss function: cross-entropy loss
    \item optimizer: Adam \cite{kingma2015adam}
    \item batch size: 32
    \item learning rate for decoder: 4e-4
    \item clip gradients at an absolute value of 5
    \item regularization parameter for ``doubly stochastic attention'': 1

\end{itemize}

The model contains 396M parameters, of which 43M are in the image encoder, 110M are in the context encoder, and the rest are in the decoder. The model has 244M trainable parameters, all of which are in the decoder, as the encoders are frozen.

\subsubsection{Compute}

All models were trained and evaluated on AWS EC2. Description-generation models were trained either on an g4dn.xlarge instance or an g5.xlarge instance. All caption-generation models were trained on an g5.xlarge instance. When training on an g4dn.xlarge instance, each model takes about 1.5 days. When training on an g5.xlarge instance, each model takes about 1 day. This includes evaluation on the validation set after each epoch, where the model generates a sequence with a beam size of 1 for each example.

For each model, evaluation on the test set takes about 2 hours on a p2.xlarge instance. Evaluation on the test set uses a beam size of 5.

\subsection{DenseNet-LSTM}

The DenseNet-LSTM follows the same structure as the ResNet-LSTM except that the ResNet-101 image encoder is replaced by a DenseNet-161. Consequently, the models differ in the number of parameters. The DenseNet-LSTM contains 382M parameters, of which 26M are in the image encoder, 110M are in the context encoder, and the rest are in the decoder. The model has 245M trainable parameters, all of which are in the decoder, as the encoders are frozen. For all other aspects, consult the ResNet-LSTM specifications.

\subsection{OSCAR(VinVL)}

\begin{figure}[tp]
  \centering
  \includegraphics[width=1\linewidth]{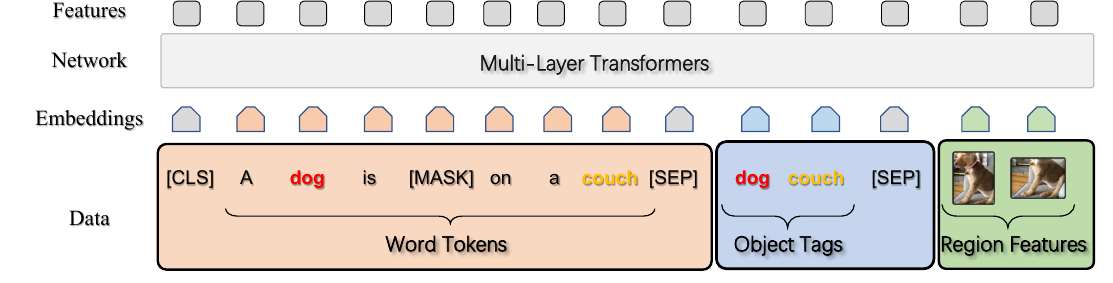}
  \caption{Illustration of the OSCAR architecture (crop taken from \citet{li2020oscar}). For each image--text pair, the image is first preprocessed by a pretrained visual feature extractor, which outputs object tags in text and region features, i.e., vector embeddings of object-enclosing regions. The OSCAR model represents the image--text pair by concatenating the text and the object tags, embedding it through BERT \cite{devlin2019berta}, then concatenating the BERT embedding with the visual features \cite{li2020oscar}. We use VinVL as the visual feature extractor \cite{zhang2021vinvl}.} 
  \label{fig:oscar-architecture}
\end{figure}

\subsubsection{Architecture}
OSCAR is a Transformer-based visual-language model that is pretrained on a large-scale corpus and can be finetuned for a variety of downstream vision-language tasks \cite{li2020oscar}, including image-to-text generation. See \figref{fig:oscar-architecture} for a diagram of the architecture, and refer to Section 4 of \citealt{li2020oscar} for the image-to-text finetuning procedure.

There are potentially many ways in which one could integrate context into the model. We took the approach of appending the context to the object tags that OSCAR receives.

\subsubsection{Implementation}
We applied the VinVL feature extractor to images in Concadia using the VinVL codebase available at \url{https://github.com/microsoft/scene_graph_benchmark}.
We adopted the OSCAR model implementation and training script from the OSCAR codebase available at \url{https://github.com/microsoft/Oscar}.

\subsubsection{Hyperparameters and Training Details}

We used the pretrained BERT\textsubscript{Base}-based OSCAR(VinVL) checkpoint as our starting point and finetuned it on Concadia. We used the same hyperparameters as the original OSCAR implementation, with two exceptions: due to compute constraints, we fine-tuned OSCAR for 20 epochs (compared to 40) and used a batch size of 32 (instead of 256). 

Aside from optimizing for cross-entropy loss, the original implementation further finetuned the model with CIDEr optimization as the objective. Since it wasn't central to our investigation, we omitted the CIDEr optimization step. By the end of training, the differences between conditions had already persisted over several epochs and validation set performance had converged.

The model contains $\approx$110M parameters, 88M of which are trainable.

\subsubsection{Compute}

All models were trained and evaluated on AWS EC2 g4dn.xlarge instances. Each model takes about 2 days to train. This includes evaluation on the validation set after each epoch, where the model generates a sequence with a beam size of 1 for each example.

For each model, evaluation on the test set takes about 8 hours on a g4dn.xlarge instance. Evaluation on the test set uses a beam size of 5.

\subsection{Validation Set Performance}

The test set performances in \tabref{tab:modelresults} are similarly reflected on \Wiki's validation set. \tabref{tab:modelresults_val} shows the validation set results. Note that evaluation on the validation set uses a beam size of 1 (different from 5 on the test set).

\begin{table}[tp]
\centering
\small
\begin{tabular}{@{} l@{ \ }l@{ \ }l@{ \ }l@{ \ }l@{ \ }@{}}\toprule
  Label & Context & RN-LSTM & DN-LSTM & OSC.VinVL \\ 
  \midrule 
Descr. & None & 0.15 (0.01) & 0.19 (0.00) & 0.24 (0.00) \\
Caption & None  & 0.07 (0.00) & 0.10 (0.01) & 0.12 (0.00) \\
\midrule
Descr. & Rand. Par. & 0.14 & 0.17 & 0.23 (0.00)\\
Caption & Rand. Par. & 0.06 & 0.08 & 0.12 (0.00)\\
\midrule
Descr. & Paragr. & 0.17 (0.01) & 0.20 (0.00) & 0.31 (0.00)  \\
Caption & Paragr. & 0.11 (0.00) & 0.13 (0.00) & 0.44 (0.00)  \\ 
\midrule
Descr. & Caption & 0.22 (0.01) & 0.24 (0.00) & 0.98 (0.01) \\
Caption & Descr. & 0.15 (0.00) & 0.17 (0.00) & 1.00 (0.00) \\ \bottomrule
\end{tabular}
\caption{CIDEr scores on validation set for all models: ResNet-LSTM, DenseNet-LSTM and OSCAR(VinVL). Standard deviations (in brackets) are computed over 3 random seeds.}
\label{tab:modelresults_val}
\end{table}

\section{Human Subject Experiment}\label{app:humaneval}

\subsection{Participants}

We recruited 421 participants over Amazon's Mechanical Turk, restricted to participation within the US and a previous average approval rate above 98\%. 
Participants were paid \$2.60 for participation with an average completion time of 12 minutes (\$13/hr).
Participants were only allowed to complete the experiment once.

\subsection{Details on Materials \& Procedure}

\begin{figure}[tp]
  \centering
  \includegraphics[width=1\linewidth]{figures/experiment-example.pdf}
  \caption{A critical trial as it appeared to the participants. Question order was randomized between participants but consistent throughout the experiment.} 
  \label{fig:exp-example}
\end{figure}

Throughout the experiment, participants never saw the same picture twice. 
Each participant saw at least seven samples of each text condition, resulting in 28 trials. The last two trial conditions were randomly sampled. An example trial as it appeared to participants is provided in \figref{fig:exp-example}. Image selection and trial order was completely randomized and question order was randomized between participants. The experiments themselves are released as part of the Github repository.

\subsection{Data Exclusions}

Participants were excluded based on their self-reported native language, their self-assessment of task performance, two attention check questions, and their overall response. Firstly, we only included participants who indicated that English is among their native languages (12 exclusions), and who reported that they thought they did the experiment correctly (74 exclusions). We constructed attention check questions where the text clearly described something different than what was displayed in the image and excluded participants who didn't select the checkbox ``Can't say because image and text seem to be unrelated'' (85 exclusions). Furthermore, we excluded participants who selected the checkbox in more than 80\% of all trials (6 exclusions). Overall, we excluded 177 participants (42\%). Furthermore, we excluded trials from the quantitative analysis where participants indicated that text and image didn't relate. If more than 80\% of participants agreed that image and text didn't relate, we excluded that data sample from our quantitative analysis as well (13,942 out of 15,616 datapoints (89.3\%) remain).

\end{document}